\documentclass[sigconf]{acmart}

\usepackage{booktabs} % For formal tables
\usepackage{multirow}

\setcopyright{rightsretained}
\acmDOI{}

\acmISBN{}

\begin{document}
\title{A Single Shot Text Detector with Scale-adaptive Anchors}
%\titlenote{Produces the permission block, and
%  copyright information}
%\subtitle{Paper ID: 390}
%\subtitlenote{The full version of the author's guide is available as
%  \texttt{acmart.pdf} document}

%\author{Qi Yuan}
%\email{yqdlut@foxmail.com}
%\author{Bingwang Zhang}
%\email{zbwdlut@foxmail.com}
%\author{Haojie Li}
%\email{lihaojieyt@gmail.com}
%\author{Zhihui Wang}
%\email{wangzhihui1017@gmail.com}
%\author{Zhongxuan Luo}
%\email{zxluo@dlut.edu.cn}
%
\author{Qi Yuan}
\affiliation{%
  \institution{}
  \streetaddress{}
}
\email{yqdlut@foxmail.com}

\author{Bingwang Zhang}
\affiliation{%
  \institution{}
  \streetaddress{}
}
\email{zbwdlut@foxmail.com}

\author{Haojie Li}
\affiliation{%
  \institution{}
  \streetaddress{}
}
\email{lihaojieyt@gmail.com}

\author{Zhihui Wang}
\affiliation{%
  \institution{}
  \streetaddress{}
}
\email{wangzhihui1017@gmail.com}

\author{Zhongxuan Luo}
\affiliation{%
  \institution{}
  \streetaddress{}
}
\email{zxluo@dlut.edu.cn}
%\affiliation{%
%  \institution{Inria Paris-Rocquencourt}
%\affiliation{%
% \institution{Rajiv Gandhi University}
%\affiliation{%
%  \institution{Tsinghua University}
%  \streetaddress{30 Shuangqing Rd}
%  \country{China}
%\affiliation{%
%  \institution{Palmer Research Laboratories}
%  \streetaddress{8600 Datapoint Drive}
%\affiliation{\institution{The Th{\o}rv{\"a}ld Group}}
%\email{jsmith@affiliation.org}
%
%\author{Julius P.~Kumquat}
%\affiliation{\institution{The Kumquat Consortium}}
%\email{jpkumquat@consortium.net}
%
%% The default list of authors is too long for headers.

\begin{abstract}
Currently, most top-performing text detection networks tend to employ fixed-size anchor boxes to guide the search for text instances. They usually rely on a large amount of anchors with different scales to discover texts in scene images, thus leading to high computational cost. In this paper, we propose an end-to-end box-based text detector with scale-adaptive anchors, which can dynamically adjust the scales of anchors according to the sizes of underlying texts by introducing an additional scale regression layer. The proposed scale-adaptive anchors allow us to use a few number of anchors to handle multi-scale texts and therefore significantly improve the computational efficiency. Moreover, compared to discrete scales used in previous methods, the learned continuous scales are more reliable, especially for small texts detection. Additionally, we propose Anchor convolution to better exploit necessary feature information by dynamically adjusting the sizes of receptive fields according to the learned scales. Extensive experiments demonstrate that the proposed detector is fast, taking only $0.28$ second per image, while outperforming most state-of-the-art methods in accuracy.
\end{abstract}




\maketitle

\section{Introduction}

In the past few years, scene text detection and recognition have received a lot of attention from both academia and industry, due to its numerous potential applications in image understanding and computer vision systems. Detecting text from natural scene is an open issue in computer vision field because texts may appear in various forms and the background may be very complex. From systematically perspective, a text detector which can detect individual words directly while being robust enough to complex background is more preferable, as it will greatly simply the processing of later recognizer \cite{DBLP:journals/corr/abs-1710-03425}.

 Owning to this, recently, many state-of-the-art text detectors \cite{Jaderberg2016,Liao2016TextBoxes,zhong2016deeptext} based on the advanced general object detection techniques \cite{NIPS2015_5638,liu2016ssd}, or box-based text detectors are proposed, which take words as the detection targets and thus make individual words detection feasible. Generally, they directly output word bounding boxes by jointly predicting text presence and coordinate offsets to anchor boxes \cite{NIPS2015_5638} at multiple scales. By this way, they have remarkably improved the detection performance, in terms of accuracy and robustness. However, we argue that the current box-based frameworks are still inefficient and unsatisfactory, for two main reasons.

First, it is not efficient to handle multi-scale texts by traversing all the possible scales (see Figure 1). The current box-based text detectors employ fixed-size anchors to match the words, and the box-regression can only adjust the sizes of anchors to some extent, the effect is rather minor. Due to the diversity of text size, they have to preset massive anchor boxes of different scales to match the underlying text shapes, which results in high computational cost. For example, in \cite{Liao2016TextBoxes} $6$ scales (implemented with $6$ layered feature maps) are used and each cell is associated with fixed-size anchors.

Secondly, it is unreasonable to match texts of all possible scales with limited discrete scaled anchor boxes. This fact has been observed in \cite{Liao2016TextBoxes,zhong2016deeptext}. In these work, though $3\sim6$ scales are adopted to produce multi-scale anchors, there are still some texts are missed when no appropriately designed scale is applicable. Therefore, fixed-size anchors have become the bottleneck for the box-based text detection framework, though they are widely adopted currently.

To conquer the above limitations, in this work, we propose a novel box-based text detector with scale-adaptive anchors, where the scales of anchors are dynamically adjusted according to the sizes of texts. Specifically, we introduce an additional scale regression layer to the basic box-based framework and use it to learn the scales of anchors in an implicit way, such that extra training supervision of object size is avoided. With the proposed scale-adaptive anchors, we only need to preset a few initial anchors of different aspect ratios in $1$ scale, thus reducing the number of anchors largely. Meanwhile, the learned scale value is continuous which is more applicable to detect various texts than several discrete scales, especially for small texts.

\begin{figure*}[t]
\includegraphics[height=3in, width=5.8in]{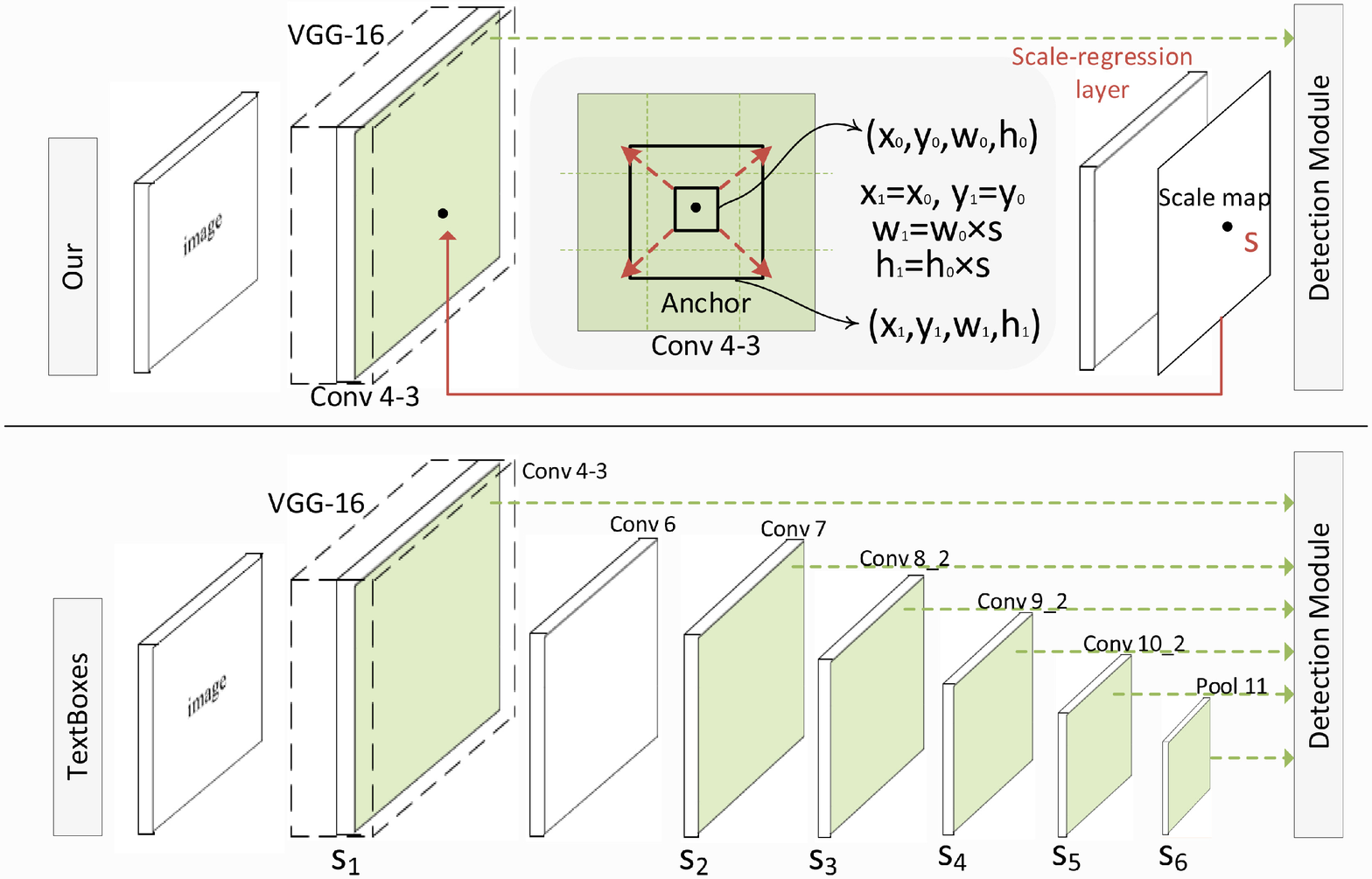}
\caption{A comparison between our model and the representative box-based models(TextBoxes)\cite{Liao2016TextBoxes}. TextBoxes add 6 convolutional feature layers(green layers) decreasing in size progressively to the end of the VGG16 model, and thus allow predictions of detections at $6$ scales. Differently, we only make predictions on $Conv 4\_3$ layer and add an additional scale regression layer behind it to predict text scale for each location of $Conv 4\_3$. Then the learnt scales are encoded to $Conv 4\_3$ layer to achieve scale-adaptive detection(the anchor boxes per location can be enlarged or shrunk according to the scale $s$ ). By using the learnt scale to replace multiple presetting discrete scales, we improve the computation efficiency(reduce the running time from 0.73s to 0.28s), while keeping competitive accuracy.
}
\end{figure*}

Additionally, we argue that when making predictions on a single feature map (\textit{i.e.}, single scale), when the scale of anchor is updating, the size of the corresponding receptive fields should change synchronously. However, for a given anchor, regardless of size, the standard convolutions in CNN \cite{NIPS2012_4824} can only assign a fixed-size respective field to it. To tackle this problem, we propose Anchor convolution to dynamically adjust the sizes of receptive fields according to the learned scales of anchors, to ensure the integrity and richness of feature information of each anchor.

To summarize, the contributions of this paper are as follows:

\begin{itemize}
 \item We propose scale-adaptive anchors which largely reduce the computational cost and improve the robustness against multi-scale texts, especially small scales. The whole framework is end-to-end, simple and easy to train.

\item We propose Anchor convolution to dynamically adjust the sizes of respective fields, to ensure the integrity and richness of feature information of each anchor.

\item We evaluate the proposed method on two real-world text detection datasets, \textit{i.e.}, ICDAR11 \cite{shahab2011icdar} and ICDAR13 \cite{karatzas2013icdar} and demonstrate that while keeping competitive accuracy with state-of-the-art, it is more efficient, taking only 0.28s per image, which is important to real systems, especially mobile applications.

\end{itemize}

\begin{figure*}[t]
\includegraphics[height=3in, width=7in]{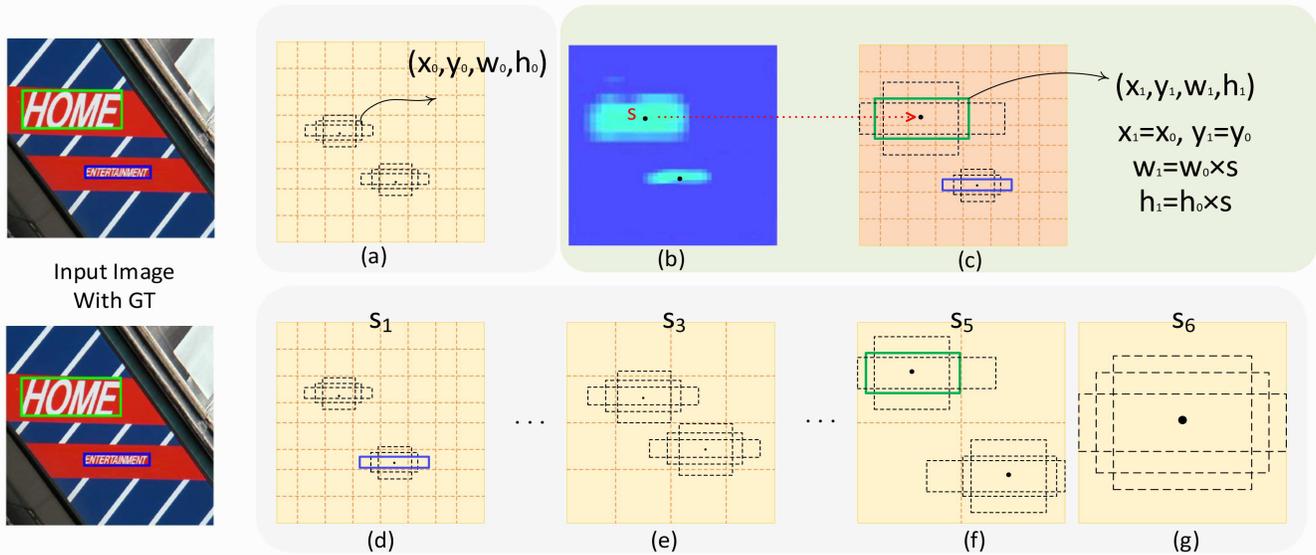}
\caption{Comparisons of our framework(Top) with TextBoxes(Bottom). (a)the initial anchors we preset on $Conv 4\_3$, and all the anchors have a initial scale.(b)generated scale map, which predicts the text size of each feature map location. (c)the sizes of initial anchors are dramatically adjusted according to the text scales learned by scale map. (d)-(g)the fixed-size anchors preseted by TextBoxes. To handle multi-scale texts, TextBoxes requires to employ 6 feature maps to produce anchors with different scales, while we only need one single map and reduce the time computational complexity from $O(n)$ to $O(1)$, the details can be seen in Sec 3.1. The green anchor and blue anchor can match the large text and small text of input image respectively.  }
\end{figure*}

\section{Related Work}

Scene text detection have been extensively studied for a few years in the computer vision community, and a large amount of methods have been proposed. Traditional methods \cite{epshtein2010detecting,huang2013text,6945320,10.1007/978-3-319-10593-2_33} deal with this issue usually by first detecting individual characters or coarse text regions, then following with a sequential processing of grouping or segmentation to form text lines or blocks. However, such post-processing steps are difficult to design because they require exploring many low-level image cues and various heuristic rules, which also make the whole system highly complicated and unreliable.

Owning to the strong representation-capability of the deep Convolutional Neural Networks (CNN), more and more deep learning based methods develop rapidly. A number of recent approaches were built on Fully Convolutional Networks \cite{Long_2015_CVPR}, by treating text detection as semantic problem. For example, Yao \textit{et al.} \cite{DBLP:journals/corr/YaoBSZZC16} propose to directly run the algorithm on the full images and produce global, pixel-wise prediction maps, in which detections are subsequently formed. Zhang \textit{et al.} \cite{Zhang_2016_CVPR} propose Text-Block FCN for generating text salient maps and the Character-Centroid FCN for predicting the centroid of characters. However, the current FCN-based methods fail to produce accurate word-level predictions with a single model, and they still require multiple bottom-up steps to construct words.

Recently, inspired by the great progress of deep learning methods \cite{NIPS2015_5638} \cite{liu2016ssd} for general object detection, many box-based text detectors are proposed and have advanced the performance of text detection considerably. For example, Jaderberg \textit{et al.} \cite{Jaderberg2016} propose an R-CNN-based \cite{7410526} framework, which first generates word candidates with a proposal generator and then adopts a convolution neural network (CNN) to refine the word bounding boxes. Liao \textit{et al.} \cite{Liao2016TextBoxes} propose an end-to-end trainable network named \textit{TextBoxes}, to directly output word bounding boxes by jointly predicting text presence and coordinate offsets to anchor boxes \cite{NIPS2015_5638} at multiple scales. However, most of them rely on presetting a large amount of anchors with different scales to discover texts in scene images, thus leading to high computational cost.

\section{Approach}
In this section, we propose a novel end-to-end text detector, which can automatically adjust the sizes of both anchors and receptive fields according to the scales of texts. The whole framework is illustrated in the top row of Figure 1. Initially, an input image is forwarded through the convolution layers of VGG16 \cite{simonyan2014very} and the $Conv 4\_3$ feature maps are produced. We add an additional scale regression layer behind the $Conv 4\_3$ feature maps to generate a scale map, which is used to indicate the text size at each location. Next, the scale map is passed to $Conv 4\_3$ layer to produce scale-adaptive anchors and flexible-size receptive fields. Finally, these anchors are classified and refined via the detection module, which contains a classification layer and a bounding box regression layer, similar to the SSD detector \cite{liu2016ssd}. The details of each phase will be described in the following.

\subsection{Scale-adaptive Anchors}
The basic idea of box-based detector is to associate a set of anchor boxes with every map location to coarsely match the ground truth texts, and then predict both classification scores and shape offsets for each anchor to obtain the final text locations. In natural scene images, texts are usually presented in various scales. To match multi-scale texts, most current box-based detectors tend to employ multiple feature maps from different levels to simultaneously make detection predictions.

For example, as shown in the bottom of Figure 1, TextBoxes \cite{Liao2016TextBoxes} adds 6 convolutional feature layers(green layers) decreasing in size progressively to the end of the VGG16 model, and thus allows predictions of detections at $6$ scales. In order to better describe this, we present its working principle in the bottom of Figure 2, where the pictures$(d)-(g)$ correspond to the green layers(from $Conv 4\_3$ to $Pool 11$) in Figure 1. We can see that different layers represent different scales, and the anchors of former layers are applied to match small scale texts. Although the way of searching all possible anchors can handle most multi-scale texts, it is obviously inefficient.

Different from previous box-based framework, we add an additional scale regression layer behind the $Conv 4\_3$ layer (as shown in Figure 1) to generate a scale map with one channel. The scale map has the same size with $Conv 4\_3$ layer and is used to encode the predicted text size for each location. Then, the scale map is used to obtain the scale-adaptive anchors.

The top row of figure 2 gives the working principle of the proposed scale-adaptive anchors. At each feature map cell of $Conv 4\_3$ layer, we place several initial anchors. Then, with the generated scale map, the anchors of each cell can be enlarged or shrunk according to the assigned scale values.

%6 anchors with different aspect ratios as the initial anchors, which can be denoted as \emph{ar}  $\in\{1,2,3,5,7,10\}$.

Specially, for a given anchor, we denote its initial size as $({x_0},{y_0},{w_0},{h_0})$, where ${x_0},{y_0}$ are the coordinates of its center, ${w_0}$ is the width and ${h_0}$ is the height. Suppose the learned scale corresponding to this anchor is $s$, then the updated anchor's size$(x',y',w',h')$ can be computed as follows:
\begin{equation}
\begin{array}{l}
x' = {x_0}, y' = {y_0}\\
w' = {w_0} \times s, h' = {h_0} \times s
\end{array}
\end{equation}
The detailed learning process will be introduced in Section 3.3.

Given an input image, previous box-based methods tend to preset all possible anchors, and the number of anchors can be represented as:
\begin{displaymath}
  {N_a} = \sum_{i=1}^{n_F} {D_i} \times {N_c}
\end{displaymath}

where $n_F$ is the number of green layers(as illustrated in Figure 1, and the $n_F$ of TextBoxes is 6), $D_i$ represents the size of $ith$ feature map(such as $Conv 4\_3$ is $38 \times 38$), and $N_c$ represents the number of anchors in each cell. $D_i$ is a constant. Besides, the anchors preset in each cell of $ith$ feature map have different aspect ratios(e.g. 1,3,5,7,10), and so $N_c$ is also a constant. Therefore, the time computational complexity of this kind of algorithms is $O(n)$.

As shown in Figure 1, we only employ one single layer to handle multi-scale texts, so the number of anchors of our detector can be represented as:
\begin{displaymath}
  {N'_a} = {D_1} \times {N_c}
\end{displaymath}

where $D_1$ represents the size of $Conv 4\_3$ feature map, the setting of $N_c$ is same with above. Therefore, our algorithm reduce the time computational complexity from $O(n)$ to $O(1)$, and thus improve the computation efficiency of algorithm. More importantly, we propose adaptive scale to replace the previous search way, which traversing all possible scales(just like exhaustive search), and allow other box-based methods to handle multi-scale texts in a more efficient way.

\subsection{Anchor Convolution}
In the previous section, we proposed a novel scale regression layer for dynamically adjusting scales of anchors. In this section, we propose Anchor convolution to dynamically adjust the sizes of respective fields and thus exploit scaled features for each anchor.

The standard convolution \cite{NIPS2012_4824} assigns a fixed-size respective field to each anchor and computes a feature vector for later classifying and box-regression. Different from standard convolution, we propose Anchor convolution to dynamically adjust the sizes of respective fields according to the scales of anchors and to extract necessary feature information for improving the performance of subsequent classification and regression. Next, we will introduce its working principle (see Figure 3).

In the convolution layers, each pixel of the output feature maps corresponds to a fixed size of receptive field ${\emph P}$ (${\emph P}$ is part of input layer, and also known as convolutional patch). Generally, a total of $\left({k_h} \times {k_w}\right)$ elements are selected from ${\emph P}$ to construct a feature vector and perform convolution operations with the kernel, where ${k_{h}}$ and $k_{w}$ is the height and width of kernel, respectively.

%%\emph{P}

In standard convolution, for each pixel, the size of corresponding ${\emph P}$ is $\left( {\left( {{k_h} - 1} \right){d_h} + 1,\left( {{k_w} - 1} \right){d_w} + 1} \right)$, here ${\emph P}$ is a rectangular respective field with the center coordinates of $\left( {{c_h},{c_w}} \right)$ and ${d_h}$, ${d_w}$ are the dilation parameters. For integer $i \in \left[ { - \left\lfloor {{{{k_h}} \mathord{\left/ {\vphantom {{{k_h}} 2}} \right. \kern-\nulldelimiterspace} 2}} \right\rfloor ,\left\lfloor {{{{k_h}} \mathord{\left/ {\vphantom {{{k_h}} 2}} \right. \kern-\nulldelimiterspace} 2}} \right\rfloor } \right]$ and integer $j \in \left[ { - \left\lfloor {{{{k_w}} \mathord{\left/ {\vphantom {{{k_w}} 2}} \right. \kern-\nulldelimiterspace} 2}} \right\rfloor ,\left\lfloor {{{{k_w}} \mathord{\left/ {\vphantom {{{k_w}} 2}} \right.
 \kern-\nulldelimiterspace} 2}} \right\rfloor } \right]$, the coordinates of selected elements from ${\emph P}$ are:
\begin{equation}
{h_{ij}} = {c_h} + i{d_h},{w_{ij}} = {c_w} + j{d_w}
\end{equation}
Let $\emph{I} = ${\emph P}$ \left( {{h_{ij}},{w_{ij}}} \right)$ denote the feature vector constructed by these elements. Then, ${\emph I}$ is used to perform element-wise multiplication with the kernel.

For Anchor convolution, suppose the output layer has the same size  with the scale map and all channels share one scale map. Thus, each pixel of the output map corresponds to a scale coefficient. Let the respective field here be ${\emph P}'$, which is also a rectangle with the same center  as ${\emph P}$. Differently, the size of ${\emph P}'$ can change to $\left( {\left( {{k_h} - 1} \right){d_h}{s} + 1,\left( {{k_w} - 1} \right){d_w}{s}\\
 + 1} \right)$ along with the scale coefficient $s$. Then, a total of ${k_{h} \times {k_{w}}}$ elements are selected from ${\emph P}'$ to construct ${\emph I}$. Inspired by \cite{Zhang_2017_ICCV}, the coordinates of these elements are changed to:
\begin{equation}
{h'_{ij}} = {c_h} + i{d_h}{s},{w'_{ij}} = {c_w} + j{d_w}{s}
\end{equation}

We adopt an irregular kernel \cite{Szegedy_2015_CVPR} in this work, setting ${k_{h}} = 1$ and ${k_{w}} = 5$. Since $i$ is integer and $i \in \left[ { - \left\lfloor {{{{k_h}} \mathord{\left/ {\vphantom {{{k_h}} 2}} \right. \kern-\nulldelimiterspace} 2}} \right\rfloor ,\left\lfloor {{{{k_h}} \mathord{\left/ {\vphantom {{{k_h}} 2}} \right. \kern-\nulldelimiterspace} 2}} \right\rfloor } \right]$, we have $i \equiv 0$ and ${h'_{ij} = {c_h}}$. In order to use scale information in height direction, we define feature vector ${\emph I}$ $ = {{\tilde {\emph P}}_{ij}}$ and construct it in a different way, which is formulated as:
\begin{equation}
\begin{array}{l}
{{\tilde {\emph P}}_{ij}} = \left( {1 - \alpha } \right){\emph P}'\left( {{c_h},{{w'}_{ij}}} \right) + \left((s>1) ? \right) \\
\cdot\left(\frac{\alpha }{2}{\emph P}'\left( {{c_h} - \frac{{{s} - 1}}{2},{{w'}_{ij}}} \right) + \frac{\alpha }{2}{\emph P}'\left( {{c_h} + \frac{{{s} - 1}}{2},{{w'}_{ij}}} \right)\right)
\end{array}
\end{equation}
where $\alpha$ is a weight parameter.

In the Anchor convolution, the respective field ${\emph P}'$ can be shrunk or expanded according to the different values of scale coefficients. As illustrated in Figure 3, when $scale = 1$, the Anchor convolution is the same with the standard convolution. However, when $scale > 1$, the respective field ${\emph P}$ will be expanded to ${\emph P}'$. In this case, we select $3\times5$ sampling elements to construct the feature vector according to Eq.4. In our work, Anchor convolution is applied in $reg$ layer and $cls$ layer.

\begin{figure}
\includegraphics[height=1.6in, width=3in]{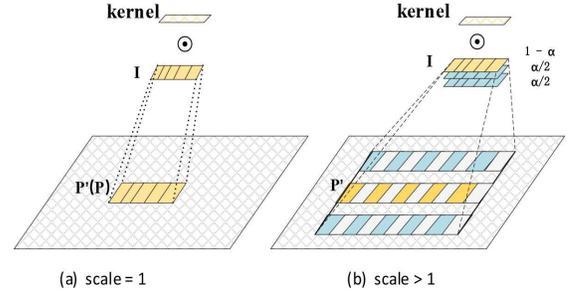}
\caption{Working principle of Anchor convolution. (a): scale = 1. (b): scale $> 1$.}
\end{figure}

\subsection{Scale Learning}

We introduce a scale regression layer to generate scale map, which is applied to scale-adaptive anchors and Anchor convolution learning. Thus, the derivations of loss function w.r.t. scale follows the chain rule.

The objective loss function is defined as follows:

\begin{displaymath}
L\left( {x,c,l,g} \right) = \frac{1}{N}\left( {{L_{conf}}\left( {x,c} \right) + \beta {L_{loc}}\left( {x,l,g} \right)} \right),
\end{displaymath}
here $x = 1$ denotes positives and $x = 0$ denotes negatives, $N$ is the number of matched anchors, $c$ is the confidence, $l$ is the predicted box, $g$ is the ground truth box, and $L_{conf}$ is a 2-class softmax loss. Smooth-L1 defined in \cite{girshick2015fast} is applied to the localization loss $L_{loc}$.

Now, we derive gradients w.r.t. scale coefficients. For brevity, we omit the standard derivations applied in network.

\textbf{In scale-adaptive anchors.} We define the predicted box as $l=\left(x,y,w,h\right)$, which is computed as: %For details, $reg$ layer produced bias $\left( {\Delta x,\Delta y,\Delta w,\Delta h} \right)$, and $l$ is calculated as:
\begin{equation}
\begin{array}{l}
x = x' + w'\Delta x\\
y = y' + h'\Delta y\\
w = w'\exp \left( {\Delta w} \right)\\
h = h'\exp \left( {\Delta h} \right)
\end{array}
\end{equation}
where $\left( {\Delta x,\Delta y,\Delta w,\Delta h} \right)$ are offsets relative to the matched anchor , which are learned from $reg$ layer. According to Eq.1, the gradient of $\left(x, y, w, h\right)$ (we use $\left(.\right)$ for brevity) w.r.t. $s$ is obtained as
\begin{equation}
\frac{{\partial \left( {.} \right)}}{{\partial  {s} }}
 = \left( {\Delta x + \exp \left( {\Delta w} \right)} \right){w_0} + \left( {\Delta y + \exp \left( {\Delta h} \right)} \right){h_0}
\end{equation}
If we use $n_{p}$ to denote the numbers of anchors at position $t$, then the gradient of $l$ w.r.t. $s_{t}$ is ${\sum\limits_{n_p}{\frac{{\partial \left( {.} \right)}}{{\partial  {s} }}}}$.

\textbf{In Anchor convolution.} We first formulate the forward propagation of our proposed Anchor convolution, and then give the formulations to update scale.

Let  $H$, $W$ and $C$ denote the height, width and channel number of feature map, respectively. We also define subscripts $in$ and $out$ to distinguish inputs and outputs. Suppose the convolution kernels in Anchor convolution is denoted as $\emph K \in \mathbb{R}^{({C_{out}}) \times ({C_{in}} \times {k_{h}} \times {k_{w}})}$ and the bias is $b \in \mathbb{R}^{C_{out}}$. We further define the subscript $x \in \left[ {1,{C_{out}}} \right]$, $y \in \left[ {1,{H_{out}} \times {W_{out}}} \right]$ and $c \in \left[ {1,{C_{in}}} \right]$. If taking $\emph K$ as a partitioned matrix, then each of its block ${\Phi_{xc}}^T \in {\mathbb{R}^{\left( {{k_h} \times {k_w}} \right)}}$ is a vector, corresponding to one of the convolution kernels. For any element ${O_{xy}}$ in convolution $output$, we have:
\begin{equation}
{O_{xy}} = \sum\limits_{c = 1}^{{C_{in}}} {{\Phi_{xc}}{{\emph I}_{cy}}}  + {b_x}
\end{equation}

In Anchor convolution, we compute the coordinates of feature vector via Eq.3. We denote the scale efficient corresponding to ${O_{xy}}$ as ${s_{y}}$.
Since $s_{y}$ is a float value, the coordinates ${w'_{ij}}$ and ${c_h} \pm \frac{{{s_y} - 1}}{2}$ may not be integers. Here, inspired by the Spatial Transformer Networks \cite{NIPS2015_5854}, we obtain ${{\tilde {\emph P}}_{ij}}$ through bilinear interpolation. Let ${{\emph I}_{cy}}$ be the feature vector corresponding to ${{\tilde {\emph P}}_{ij}}$, the forward propagation of convolution is computed via Eq.7.

During backward propagation, after obtaining $g\left(O_{xy}\right)$ from $loss$ layer, the gradients w.r.t. ${\emph I}_{cy}$, $ \Phi_{xc}$ and $b_{x}$ are derived as
\begin{eqnarray}
g\left( {{{\emph I}_{cy}}} \right) &=& \sum\limits_x {\Phi _{xc}^Tg\left( {{O_{xy}}} \right)} \\
g\left( {{\Phi _{xc}}} \right) &=& \sum\limits_y {g\left( {{O_{xy}}} \right){\emph I}_{cy}^T} \\
g\left( {{b_x}} \right) &=& \sum\limits_y {g\left( {{O_{xy}}} \right)}
\end{eqnarray}

According to Eq.3, we can obtain the gradients: $\frac{{\partial {{\emph I}_{cy}}}}{{\partial {{h'}_{ij}}}}$ and $\frac{{\partial {{\emph I}_{cy}}}}{{\partial {{w'}_{ij}}}}$. Since the coordinates ${{h'}_{ij}}$ and ${{w'}_{ij}}$ rely on the scale coefficient $s_{y}$, to obtain the gradient of $s_{y}$,we first compute the partial derivatives of coordinates as follows.:\begin{equation}
\frac{{\partial {{h'}_{ij}}}}{{\partial {s_y}}} = 0 ~ or ~ \frac{\alpha }{2} ~ or ~  - \frac{\alpha }{2}, ~ \frac{{\partial {{w'}_{ij}}}}{{\partial s}} = j{d_w}
\end{equation}

Thus the final gradients of $s_{y}$ are obtained as
\begin{equation}
g\left( {{s_y}} \right) = \sum\limits_c {\sum\limits_{i,j} {{{\left( {\frac{{\partial {{h'}_{ij}}}}{{\partial {s_y}}}\frac{{\partial {{\emph I}_{cy}}}}{{\partial {{h'}_{ij}}}} + \frac{{\partial {{w'}_{ij}}}}{{\partial {s_y}}}\frac{{\partial {{\emph I}_{cy}}}}{{\partial {{w'}_{ij}}}}} \right)}^T}} g\left( {{{\emph I}_{cy}}} \right)}
\end{equation}

According to Eq.6 and Eq.12, the gradients of scale coefficients can be automatically calculated from the gradients of the following layers. In other words, the scale map can be obtained in a data-driven manner and we do not need any extra supervision. All above the derived formulations can be computed efficiently and implemented in parallel on GPUs. In practice, we limit the scale coefficients greater than zero and smaller than the size of image.

\section{Experiments}
%In this section, we first do ablation studies on synthetic data to evaluate our proposed ROI convolutions. Then we  . Finally, we apply our text detector on scene text benchmark.
We implement the proposed algorithm with Caffe \cite{Jia:2014:CCA:2647868.2654889} on Python. All the experiments are conducted on a regular server (3.3GHz 20-core CPU, 64GB RAM, NVIDIA TITAN GPU and Linux 64-bit OS) and the routine run on a single GPU in each time.

\subsection{Datasets and Experimental Settings}

{\bfseries VGG SynthText-Part.} The VGG SynthText datasets \cite{gupta2016synthetic} consist of approximately $800k$ synthetic scene-text images. For efficiency, we randomly select $500k$ images for training and refer it as VGG SynthText-Part.
%and 3000 images for exploiting alternative settings,

{\bfseries ICDAR13.} The ICDAR13 datasets are from the ICDAR 2013 Robust Reading Competition \cite{karatzas2013icdar}, with $229$ natural images for training and $233$ images for testing.

{\bfseries ICDAR11.} The ICDAR11 datasets are from the ICDAR 2011 Robust Reading Competition \cite{shahab2011icdar}, with $229$ natural images for training and $255$ images for testing.

%We used the pre-trained VGG-16 \cite{simonyan2014very} as the initialization of our network. We first train our model on VGG SynthText-Part for 20\textit{k} iterations, which includes $500k$ synthesized text images. Then we finetune it on ICDAR13 using SGD with learning rate $10^{-4}$ , $0.9$ momentum, and $5 \times 10^{-4}$ weight decay for 2\textit{k} iterations.

Our model is trained with $300 \times 300$ images using stochastic gradient descent (SGD). Momentum and weight decay are set to $0.9$ and $5 \times 10^{-4}$ respectively. Learning rate is initially set to $10^{-3}$, and decayed to $10^{-4}$ after $20k$ training iterations. We first train our model on VGG SynthText-Part for 40\textit{k} iterations, and then finetune it on ICDAR13 for $2k$ iterations. 

Compared to previous box-based methods, the number of anchor boxes in our algorithm are largely reduced, so we employ all anchors without negative mining for training. Accordingly, we add a balance parameter in our loss function to balance the ratio of positive and negative anchors. To further boost detection recall, we rescale input image to 6 resolutions, and the total running time is 0.28s.

\subsection{Visualization and Analysis}

To verify the ability of our network in learning scales of texts, we visualize the training results of scale maps after different iterations. As shown in Figure 4, the scale maps exhibit overall similar structures to the texts in images gradually when the iteration increases. Specifically, large texts have large scale values, whereas small texts have small scale values. Besides, for each text area, the scale values associated with the center points are slightly larger, while the scale values close to boundaries are slightly smaller.

\begin{figure}[h]
\includegraphics[height=1.6in, width=3in]{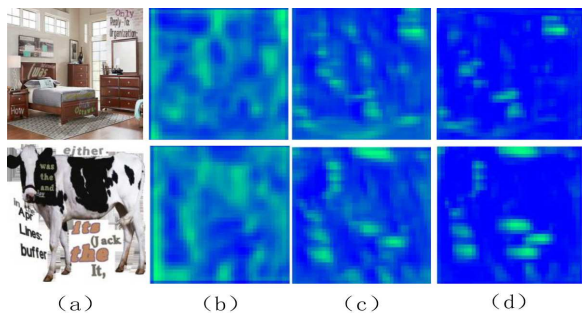}
\caption{Visualization of scale maps training results after different iterations. Different color (in green) brightness denotes different levels of scales, and brighter color represents larger value. (a)Input images; (b)500 iterations; (c)5,000 iterations; (d)30,000 iterations.}
\end{figure}

\subsection{Evaluation on Anchor Convolution}
In this part, we investigate the effect of Anchor convolution, which is used to adjust the size of receptive field of each anchor and get more abundant feature information. Two models are trained using all $50k$ images of SynthText-Part and refined on ICDAR13. One model is with Anchor convolution (denoted as AC-model) while the other is not (denoted as WAC-model). We evaluate two models on ICDAR13 and the results are tabulated in Table 1. Here P, R and F are abbreviations for Precision, Recall, and F-measure respectively.

\begin{figure*}[t]
\includegraphics[height=3in, width=6in]{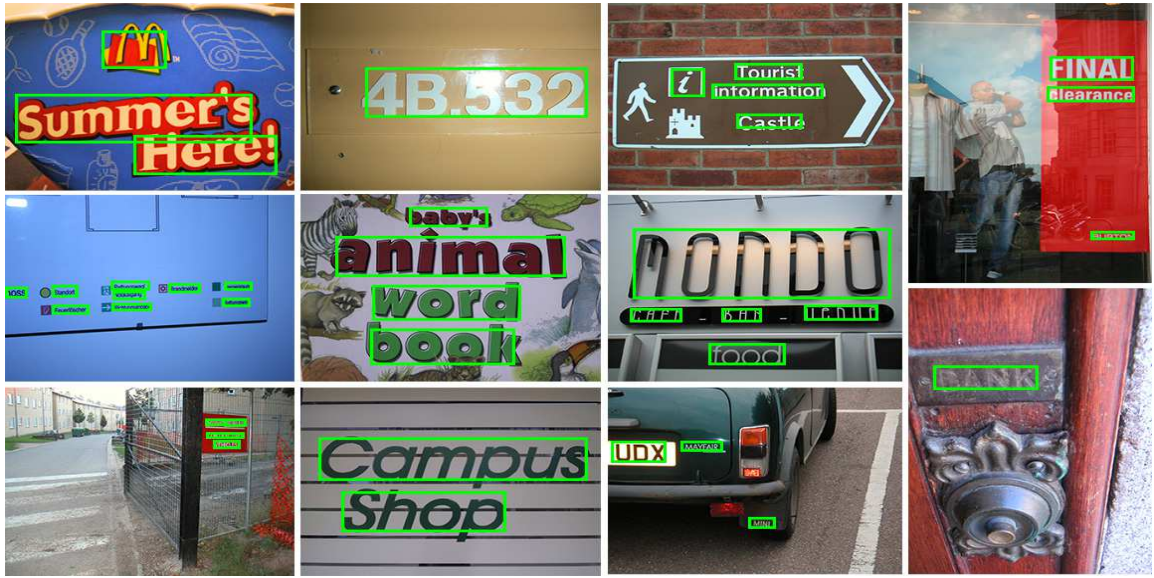}
\caption{Our detection results on several challenging images. The green bounding boxes are correct detections.}
\end{figure*}

\begin{table}[h]
  \caption{Impact of Anchor convolution on text detection. $\Delta F$ is the improvement of AC-model over WAC-model. }
  \label{tab:freq}
  \begin{tabular}{ccccl}
    \toprule
    Model  & P &    R     &    F   &  $\Delta F$  \\
    \midrule
    WAC-model         &  77\%& 76\% &  76\% & -       \\
    AC-model          &  89\%& 83\% &  86\%  & 10\% \\
  \bottomrule
\end{tabular}
\end{table}

From Table 1 we can see  that the F-measure based on Anchor convolution (AC-model) is $86\%$, which has $10\%$ improvement over WAC-model. This verifies that Anchor convolution is effective in exploiting necessary feature information by adjusting the receptive fields dynamically, for detecting texts of various sizes.

\subsection{Evaluation for Full Text Detection }

\begin{table*}
  \caption{Experimental results on the ICDAR11 and ICDAR13 datasets}
  \label{tab:commands}
  \begin{tabular}{cccccccl}
    \toprule
    Datasets     & \multicolumn{3}{|c|}{ICDAR11} & \multicolumn{3}{|c|}{ICDAR13} & \multirow{2}*{Runtime/s} \\
    \cline{1-7}
    Methods     &                                                     P     &     R     &       F        &      P     &     R     &       F     &               \\
    \midrule
    TextBoxes \cite{Liao2016TextBoxes}                           &      88\%   &    82\%     &         85\%     &      88\%      &      83\%    &      85\%  &     0.73          \\
    Yao \textit{et al.}\cite{DBLP:journals/corr/YaoBSZZC16}                                          &       -   &    -      &         -         &      89\% & 80\%  &  84\%    &         0.62      \\
    MCLAB\_FCN \cite{Zhang_2016_CVPR}            &       -   &    -      &         -         &      78\% & 88\%  &  82\%    &         2.1      \\
    RRPN \cite{DBLP:journals/corr/MaSYWWZX17}                   &       -   &    -      &         -              &      90\%       &       72\%  &  80\%       &     -          \\
    TextFlow  \cite{Tian:2015:TFU:2919332.2920067}               &      86\%  &      76\%  &          81\%   &    85\%        &   76\%      &   80\%      &     1.4          \\
    Lu \textit{et al.} \cite{Lu2015}                       &      -    &     -     &            -   &      89\%       &   70\%      &   80\%      &          -    \\
    Neumann \textit{et al.} \cite{neumann2015efficient}         &       -   &    -      &         -      &      82\%      &    72\%     &   77\%       &      0.8         \\
    FASText \cite{Buta:2015:FEU:2919332.2919945}                            &    -     &           &        -       &     84\%       &    69\%     &     75\%    &        0.55         \\
    Ours                                                         &         89\% &          82\% &               \textbf{85 \%} &         89\%  &            83\%  &  \textbf{86\%}    &    \textbf{0.28}             \\
    \bottomrule
  \end{tabular}
\end{table*}

We evaluate our detector on two benchmarks: ICDAR11 and ICDAR13. The comparison results with some state-of-the-art methods, including traditional methods and box-based methods, are tabulated in Table 2. We can see that our method achieves slightly superior results of $85\%$, $86\%$ F-measure to state-of-the-art approaches, while costing much less time, which is important to real systems especially mobile applications.

Some detection examples are given in Figure 5. The results show that our model is extremely robust against multiple text variations, cluttered backgrounds and challenging conditions like high light, blurring and so on.

\begin{figure}
\includegraphics[height=2in, width=3in]{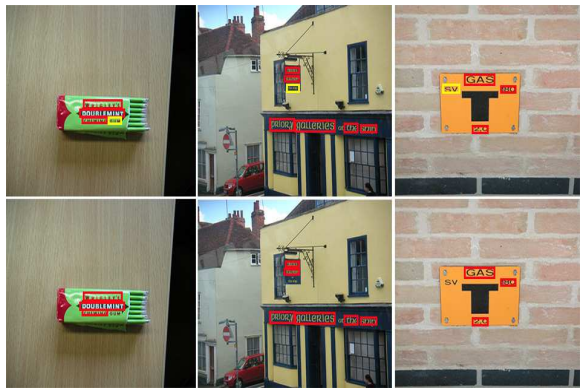}
\caption{Comparisons of our method (Top row) with TextBoxes (Bottom row). The red bounding boxes are the detection results. Our method is more efficient to handle small texts, as marked by yellow bounding boxes. }
\end{figure}

{\bfseries Advantages on Small Texts.}
We argue that our method is superior in detecting small texts. To verify this, we compare it with the representative box-based method, i.e., \textit{TextB\\oxes} \cite{Liao2016TextBoxes}. To detect small texts, \textit{TextBoxes} needs to resize the input image to $1600 \times 1600$ pixels before sending into the network, which is very time-consuming. In contrast, we can cover most of the small texts with only $800 \times 800$ input images. With such settings, as shown in Figure 6, our model is more reliable and finds all the small texts, while \textit{TextBoxes} has missed some of them. We attribute the advantages of our method on small texts to the scale-adaptive anchors and Anchor convolution.

First, different to the fixed-size anchors of several discrete scales used in \textit{TextBoxes}, the proposed scale-adaptive anchors can change their sizes continuously and thus are more potential in matching the shapes of small texts. Moreover, in \textit{TextBoxes} even the smallest anchor may be much bigger than the small texts for some test images with its current settings. Second, the proposed Anchor convolution is able to shrink the respective fields of small texts adaptively, therefore we can focus on texts while remove the side-effect of background in feature extraction.

{\bfseries Performance Analysis.} By producing scale-adaptive anchors to replace presetting all possible anchors of different scales, which are employed in most box-based methods, we improve the computation efficiency(reduce the time computational complexity from $O(n)$ to $O(1)$, the details can be seen in Sec 3.1), and reduce the running time from 0.73s to 0.28s while keeping competitive accuracy. Our running time includes generating scale map and matching anchors. Therefore, the savings in time will be more significant as the networks go deeper. Furthermore, the proposed adaptive scale allows other box-based methods to handle multi-scale texts in a more efficient way and further improve their performance.

\section{Conclusions}

In this paper, we have presented an end-to-end text detector with scale-adaptive anchors. It can largely reduce the number of anchors and thus improve the computation efficiency. Meanwhile, it also eliminates the unreliability of detection caused by discrete scales, and is more effective to handle multi-scale texts, especially small texts. Additionally, the Anchor convolution is proposed to further improve detection performance via exploiting necessary feature for each anchor. Furthermore, the proposed adaptive scale can also be applied to other methods, and allow them to handle multi-scale texts in a more efficient way. Experimental results show that our approach is fast while maintaining high accuracy on ICDAR11 and ICADR13. In future, we are interested to apply our method to arbitrary-oriented text detection task.

\begin{acks}
The authors would like to thank the associate editor and the anonymous reviewers for their constructive suggestions.
  %The authors would like to thank Dr. Yuhua Li for providing the
%  MATLAB code of the \textit{BEPS} method.
%
%  The authors would also like to thank the anonymous referees for
%  their valuable comments and helpful suggestions. The work is
%  supported by the \grantsponsor{GS501100001809}{National Natural
%    Science Foundation of
%    China}{http://dx.doi.org/10.13039/501100001809} under Grant
%  No.:~\grantnum{GS501100001809}{61273304}
%  and~\grantnum[http://www.nnsf.cn/youngscientists]{GS501100001809}{Young
%    Scientists' Support Program}.

\end{acks}

\bibliographystyle{ACM-Reference-Format}
\bibliography{sample-bibliography}

\end{document}